\documentclass{article}

\usepackage{arxiv}

\usepackage[utf8]{inputenc} 
\usepackage[T1]{fontenc}    
\usepackage{booktabs}       
\usepackage{microtype}      
\usepackage{graphicx}
\usepackage{natbib}
\usepackage{doi}

\usepackage{subfigure}
\usepackage{comment}
\usepackage{relsize}
\usepackage[english]{babel}
\usepackage[parfill]{parskip}
\usepackage{float}
\usepackage{amsmath,amssymb,mathtools,bm,etoolbox}
\usepackage[normalem]{ulem}
\usepackage{xcolor}
\usepackage{breqn}

\title{Leveraging in-domain supervision for unsupervised image-to-image translation tasks via multi-stream generators}

\author{{\hspace{1mm}Dvir Yerushalmi}\\
    Tel-Aviv University \\
	\texttt{dviry@mail.tau.ac.il} \\
	\And
	{\hspace{1mm}Dov Danon}\\
    Tel-Aviv University \\
    \And
    {\hspace{1mm}Amit H. Bermano}\\
    Tel-Aviv University \\
}

\date{}

\renewcommand{\shorttitle}{\textit{arXiv} Template}

\hypersetup{
pdftitle={Leveraging in-domain supervision for unsupervised image-to-image translation tasks via multi-stream generators},
pdfsubject={cs.LG},
pdfauthor={Dvir Yerushalmi, Dov Danon, Amit H. Bermano},
pdfkeywords={Multi-stream generators, GAN},
}

\begin{document}
\maketitle
\begin{abstract}
	Supervision for image-to-image translation (I2I) tasks is hard to come by, but bears significant effect on the resulting quality. In this paper, we observe that for many Unsupervised I2I (UI2I) scenarios, one domain is more familiar than the other, and offers in-domain prior knowledge, such as semantic segmentation. We argue that for complex scenes, figuring out the semantic structure of the domain is hard, especially with no supervision, but is an important part of a successful I2I operation. We hence introduce two techniques to incorporate this invaluable in-domain prior knowledge for the benefit of translation quality: through a novel Multi-Stream generator architecture, and through a semantic segmentation-based regularization loss term. In essence, we propose splitting the input data according to semantic masks, explicitly guiding the network to different behavior for the different regions of the image. In addition, we propose training a semantic segmentation network along with the translation task, and to leverage this output as a loss term that improves robustness. We validate our approach on urban data, demonstrating superior quality in the challenging UI2I tasks of converting day images to night ones. In addition, we also demonstrate how reinforcing the target dataset with our augmented images improves the training of downstream tasks such as the classical detection one.
\end{abstract}


\section{Introduction}
\label{sec:introduction}
Image-to-image translation is one of the most exciting new capabilities emerged by modern deep learning-based image processing approaches. Typically for this task, a source image from domain $A$ is augmented such that it holds the distribution of another domain $B$, but still maintains its original semantic structure \cite{pix2pix2017}. This task is intuitive, and yields the best results, when full supervision is available, i.e., when pairs of images from the source domain $A$ and their corresponding counterparts from the target domain $B$ are available during training. Unfortunately, this type of data is usually unavailable, or very hard to collect. 

Therefore, state-of-the-art methods in Generative Adversarial Networks (GANs) are available that perform Unsupervised Image-to-Image (UI2I) translations, i.e. learn a mapping from one image domain to another using unpaired images \cite{zhu2017toward}. While these approaches offer training processes with attainable data, they often suffer from artifacts and low visual quality \cite{unit}. This phenomenon is not surprising, and stems from several reasons. Probably one of the main ones is that figuring out the semantic structure that should be maintained during the translation operation is extremely hard, especially with no supervision, and even more so when images portray complex scenes of many entities.

In this work, we observe that a common case for practical uses of image-to-image translation indeed requires the use of unpaired data between the domains, but also offers \textit{in-domain supervision}; For many typical cases, the translation is performed between a domain that is easier in terms of data generation, collection, and/or labeling. WLOG, assume this domain is the source one, $A$, and the other is the target one, $B$. For example, in the common case of the cg2real task \cite{cg2real}, the synthetic domain can hold full labeling at virtually no added cost, and is indeed typically considered to be the source domain. Under this assumption, the source domain offers supervision regarding its semantic structure that can, and should, be leveraged. Equipped with this insight, we examine in this paper how the existence of a reliable semantic segmentation extraction technique for the source domain alone (e.g, a network trained with full supervision) benefits the task of UI2I. We introduce two techniques to incorporate the invaluable in-domain prior knowledge: one through a novel multi-stream generator architecture, and another through a semantic segmentation-based regularization loss term.

\begin{figure*}[t]
    \centering
    \includegraphics[width=.99\textwidth]{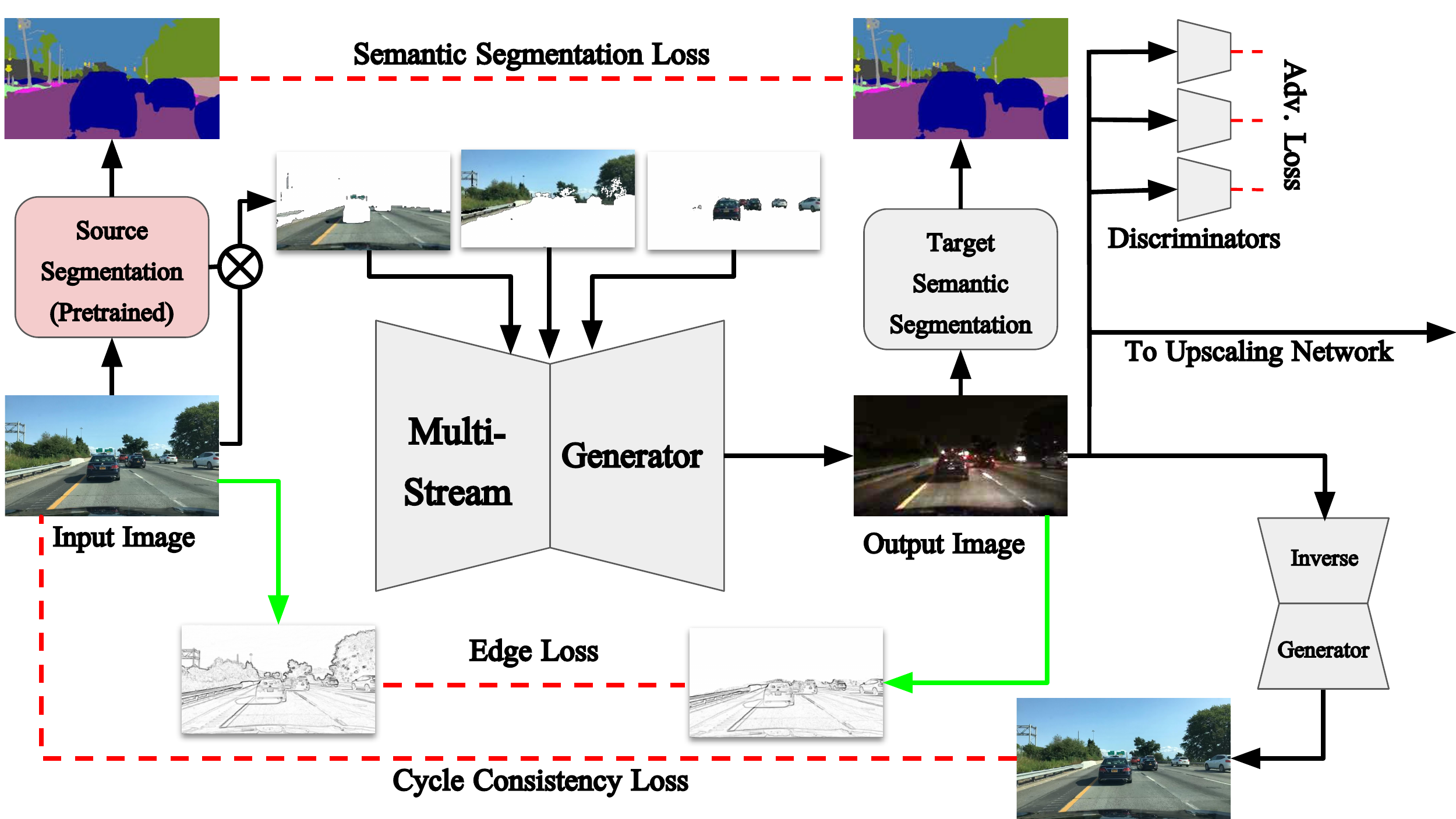}
    \caption{Multistream framework overview}
    \label{fig:Framework}
\end{figure*}

Our loss term is based on the observation that if the semantic structure of an image is indeed preserved during the translation operation, each semantic region of the source image should still be recognizable after the translation. Therefore, we propose training a semantic segmentation network along with the translation generator, that seeks to produce a segmentation map from an image in the target domain $B$. By penalizing the whole system according to the quality of the produced segmentation, the generator is encouraged to generate images in which the various elements in the target image are easily distinguishable. Note that this is not the same as asking the network to produce a segmentation map along with the translated image --- an approach that has been attempted before. For more detail regarding the fundamental difference between the two approaches, and for experimental comparisons, please see the Supplementary Material. In conjunction with traditional loss terms such as cycle consistency, perceptual, or reconstruction losses, this term assumes a regularizing role, helping with mode collapse issues, and other artifacts that make the various regions of the generated image less understandable. As a side product, this process also yields a semantic segmentation network for the target domain. However, we have seen that training a rather weak, low parameter-count architecture for this semantic segmentation part actually offers better regularization and hence induces better results. Hence, the resulting network is not recommended to be used for stand-alone semantic segmentation needs.

\begin{figure}[h]
    \centering
    \includegraphics[width=0.99\columnwidth]{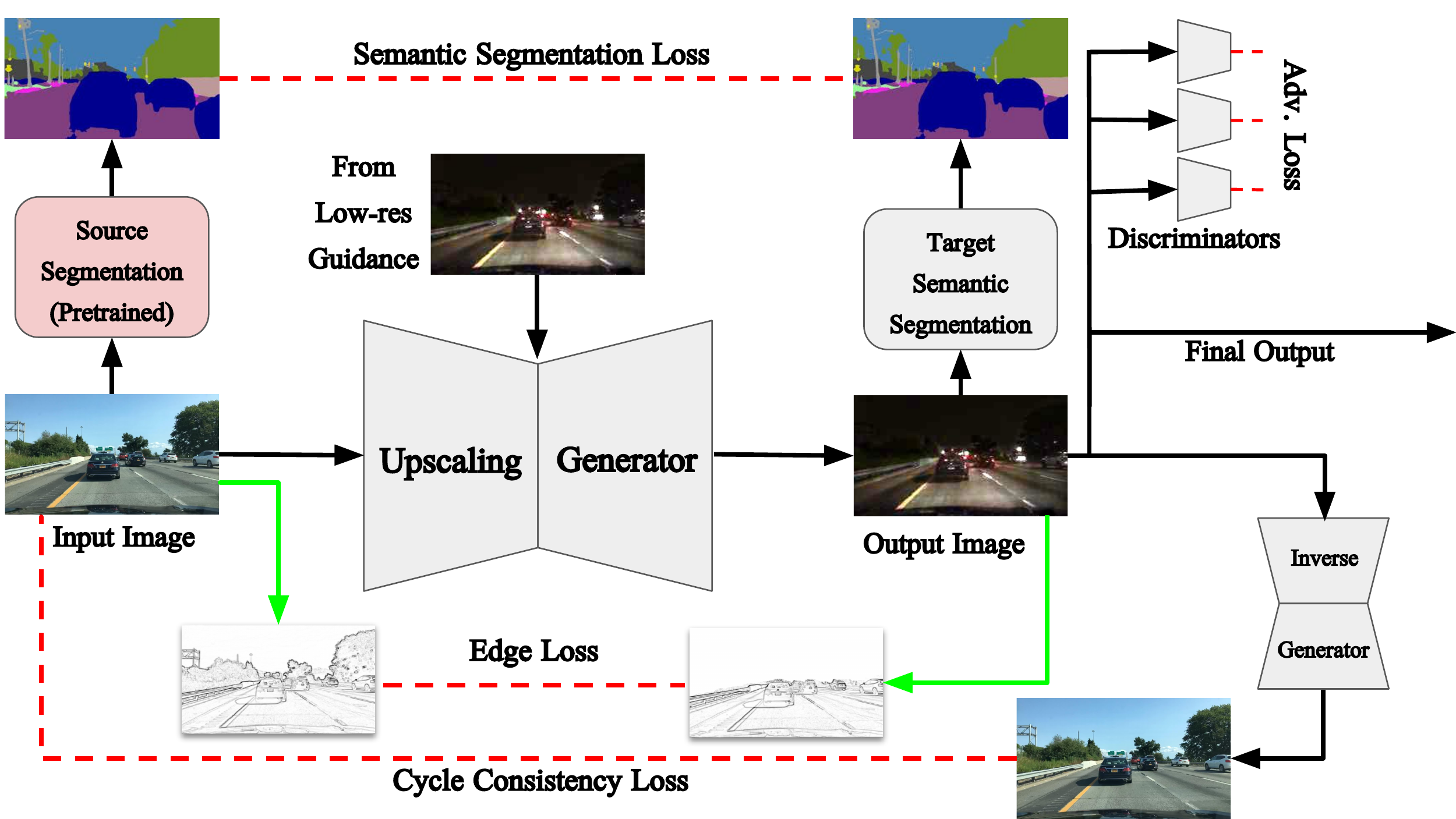}
    \caption{Upscaling framework overview}
    \label{fig:upscale}
\end{figure}

Our novel generator architecture arises from the observation that typically the translation task is multi-modal, i.e., different semantic regions in the image present different translation distributions. For example, when translating day images to night ones, white stripes on the road typically shine as they reflect headlight illumination, while white stripes on buildings do not. In accordance with our main observation, it is difficult for a training process to realize this multi-modal behavior, and properly assign the different translation behavior to different image regions during inference. Hence we propose designing dedicated pathways, or \textit{streams}, in the network for predefined semantic regions. In the aforementioned case, this means that the regions of the image that portray cars are fed through a different route in the generator than those of roads or buildings. We implement this notion by generating masks from the semantic segmentation data, and feeding different parts of the image to these different streams. Since information passing between the regions is also important, the different streams are merged in the depths of the generator before the final image is produced. as we demonstrate in Section \ref{sec:results}, this semantic separation on the architectural level indeed induces different, and more correct, behavior for each of its streams. Again, note that this architecture provides stronger guidance, and separation, than feeding the different parts of the image as different channels of one path. We demonstrate how this difference effects output quality in the Supplementary Material.

We validate our observations for complex scenes on urban data, performing two translation tasks - from day to night and from synthetic urban images to real ones. In addition, we show how using the translated images as a mean of augmentation, we improve accuracy for downstream tasks, such as detection.

Our main contributions in this paper are:
\begin{itemize}
  \item Enforcing a strong semantic segmentation prior onto the architecture itself to achieve specialized Streams within the generator (Multi-Stream architecture)
  \item Training a semantic segmentation network as an integral part of the framework to employ a semantic segmentation loss between the input and output images (non pre-trained)
\end{itemize}


\section{Related work}
\label{sec:Relatedwork}

Recent generative models, such as Generative Adversarial Networks (GANs) \cite{goodfellowGAN}, are powerful enough to achieve satisfactory results in many computer vision tasks, such as, image generation, image editing and image inpaiting \cite{shepard, Karras_2020_CVPR, abdal2020styleflow, collins2020editing}.

When guidance regarding the desired end result is needed, a Conditional GANs can be employed, these model can use various types of information, such as, images, text or labels to guide the image generation process \cite{cgan,singlepair,liu2020diverse}.
An important example for such conditional task is image-to-image translation, where the model learns a translation function to translate an image from one domain to another.

It is sometimes noted that during the translation process, we would like to preserve the content, but change the style.
For example, translate a daytime image to a nighttime one, such as the scene (objects and their composition) is preserved, but the time of day is night \cite{crossnet}.
Some works \cite{huang2018munit} tried to explicitly separate between the content and style components of each image.
Other works \cite{liu2019few} showed Few-Shot Generalization - transferring to to a domain which was not observed during training, using a only a few examples.
The task of image-to-image translation can be separated into two categories, the first is the supervised image-to-image translation, and the second is the weakly-unsupervised image-to-image translation.
The supervised \cite{pix2pix2017,wang2018pix2pixHD,park2019SPADE} category consists of paired samples in the data, i.e., the original image and its translated image.
The unsupervised \cite{disco,CycleGAN2017,DRIT,choi2018stargan,MSGAN,liu2019gesture,Kim2020U-GAT-IT:, patashnik2020balagan} category does not contain pairings, but two (or more) image domains, unsupervised/unpaired image-to-image translation aims
at learning a conditional image generation function that
can map an input image of a source class to an analogues image of a target class without pair supervision.

Some works \cite{Anokhin_2020_CVPR} employ a semantic segmentation loss where the input image has a ground truth segmentation map, and the generator is tasked with predicting this map as an extra output, on top of the translated image.
This differs from out method where a semantic segmentation discriminator is predicting a semantic map on the \textit{output} image itself.
In our method, the preservation of the semantic structure of the output image is enforced explicitly, where there is no such guarantee in the other methods.

Moreover, some works \cite{Ashual_2019_ICCV} make use of semantic segmentation maps of the input image as prior knowledge to the generator, however, they use this information to encode entity labels for the decoder and do not compel the generator into having specialized streams for specific entities.


\section{Our approach/method}
\label{sec:method}

We present a framework for the unsupervised image-to-image translation task, which leverages in-domain understanding for complex yet well examined scenes, such as urban ones. Our framework is centered around a backbone generator, and offers alterations for it to adapt for the nature of the data, as depicted in Figures~\ref{fig:Framework} and \ref{fig:upscale}. As the most established, and simple, architecture for unsupervised image-to-image translation, we demonstrate our framework using the CycleGAN network~\citep{CycleGAN2017}, however this back-bone could be replaced with another, as long as it is adapted accord to the observations and alterations presented in this section:
The task of Image-to-Image translation of urban scenes is challenging since the scenes are diverse on one hand, depicting multiple objects of different kinds, but also contain global dependencies, such as reflections from light sources, especially at night time. In addition, urban scenes also typically consist of objects of very different appearances and vastly different textures, from smooth and specular cars, through flashy billboards, to rough and diffuse walls and roads.

\begin{figure}[H]
    \centering
    \includegraphics[width=.99\columnwidth]{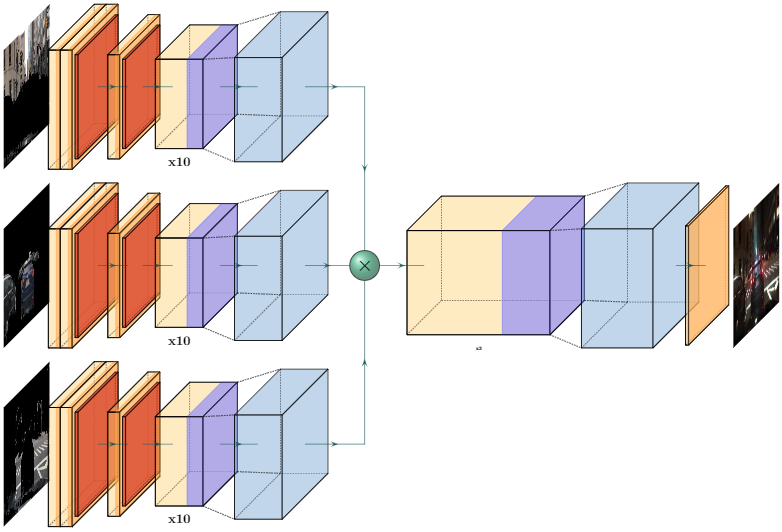}
    \caption{Main Multi-stream generator architecture}
    \label{fig:MainMultistream}
\end{figure}

\begin{figure}[H]
    \centering
    \includegraphics[width=.99\columnwidth]{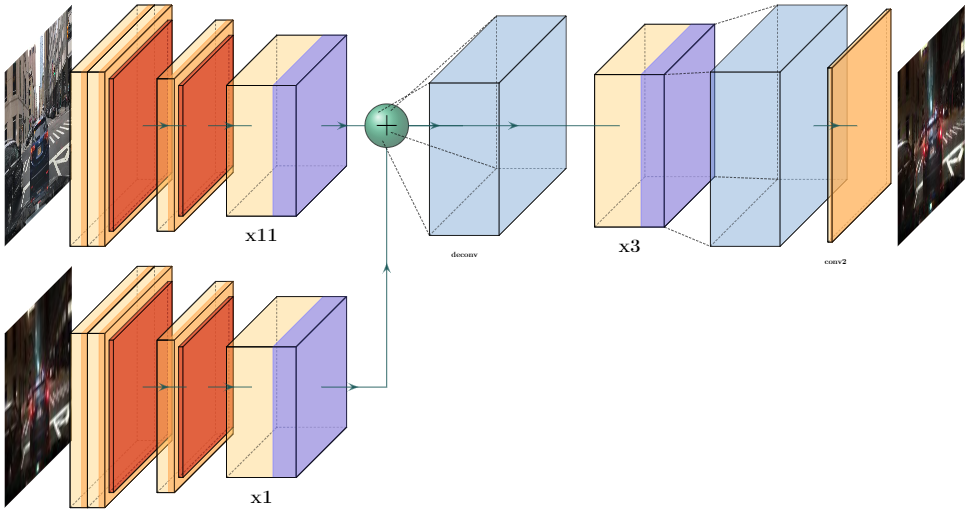}
    \caption{Up-sampling Multi-stream generator architecture}
    \label{fig:Upsampling}
\end{figure}

\begin{figure*}[h!]
    \centering
    \includegraphics[width=16cm]{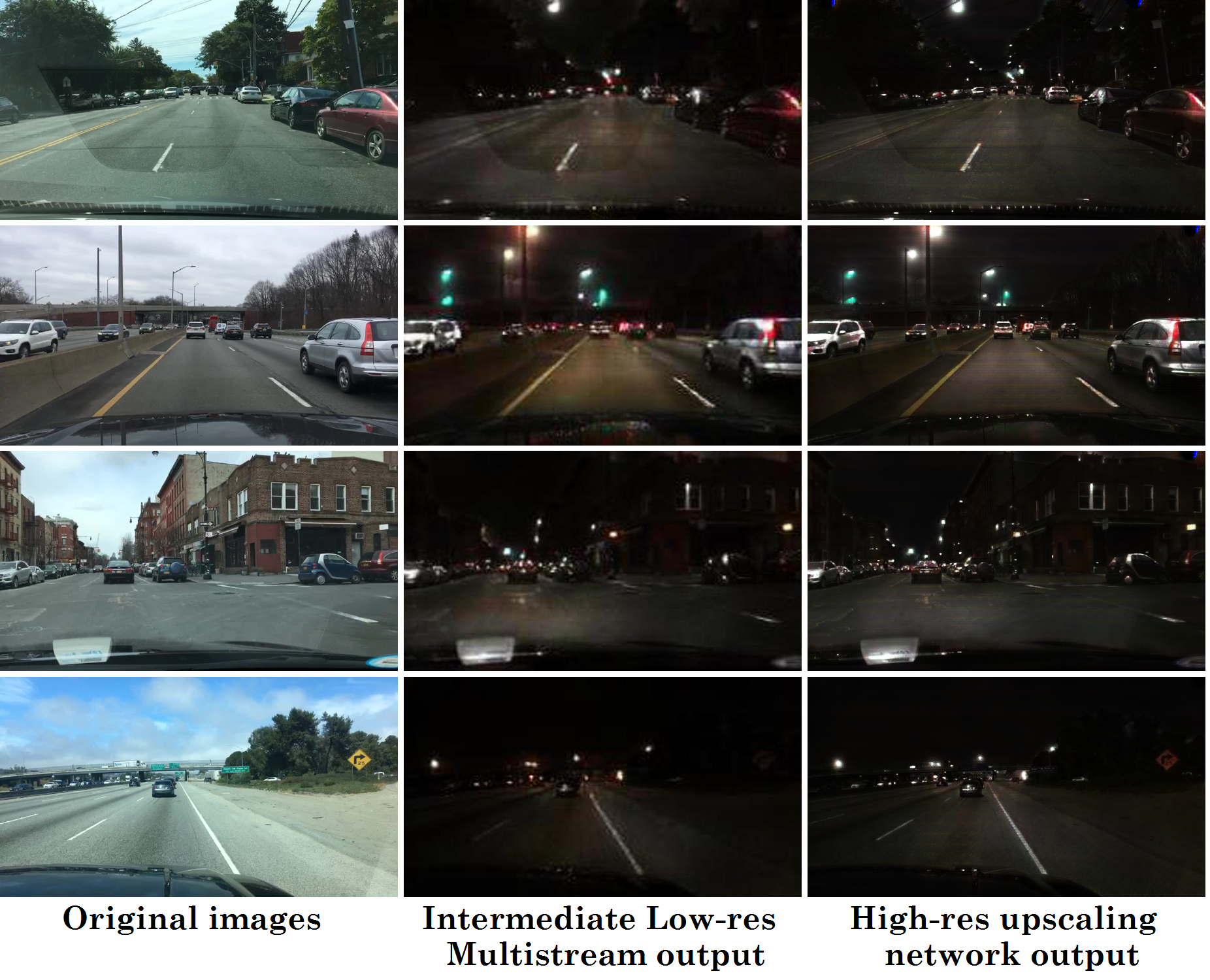}
    \caption{Intermediate guidance result: we show the translation guidance image (middle) translated by the low-res Multi-stream network. This intermediate guidance image is fed to the Upscaling network, which adds fine details, to produce the final result. Zoomed-in viewing required.}
    \label{fig:lowres}
\end{figure*}

\begin{figure*}[h!]
    \centering
    \includegraphics[width=16cm]{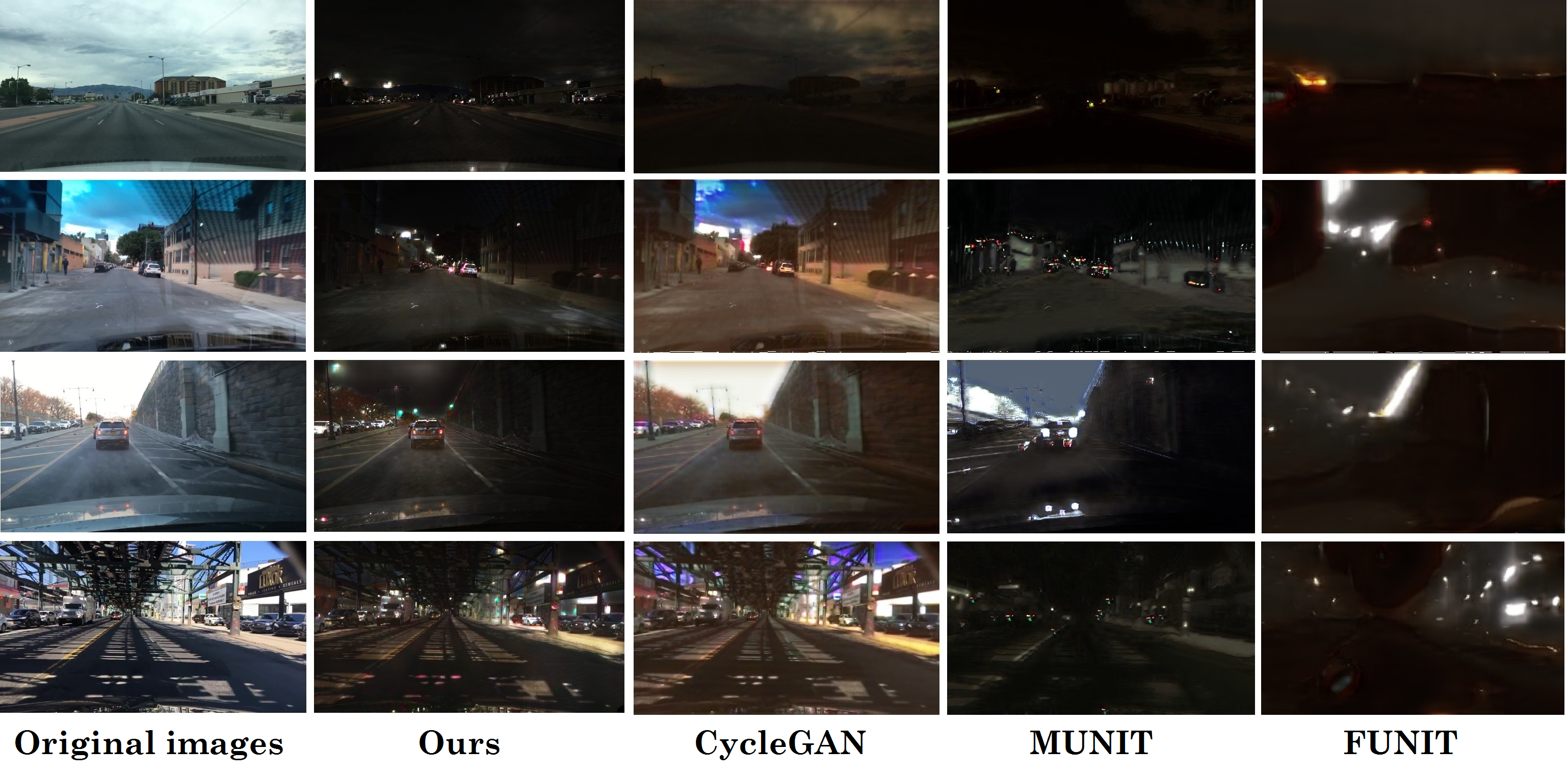}
    \caption{Day-to-night translation comparison. As can be seen, baseline methods suffer from partial translation (e.g. daytime appearance for the CycleGAN result) and misplaced visual artifacts (e.g. additional light sources for the FUNIT result). Zoomed-in viewing recommended.}
    \label{fig:comparisons}
\end{figure*}

Due to the challenging nature of these scenes, we have found that the network, and more specifically the training process, must be heavily guided in order to produce high quality results. For example, a common phenomenon in the day-to-night translation is that uninformed networks tend to scatter multiple light-sources throughout the scene, much more than the real scene should actually contain (see Figure~\ref{fig:comparisons}). This phenomenon probably stems from how prominent light-sources are in night scenes, making their saliency outweigh their structural context. Hence, the network and training scheme we introduce are designed to tackle the problems that are typical to this type of data. We propose performing four adaptations to the backbone architecture, each addressing a different problem.

\subsection{Multi-Stream Generator}
\label{sec:multistream}

First, we observe that due to all the aforementioned reasons, the translation is different for each type of object in the scene. Hence, informing the network of the semantic meaning of image regions provides powerful guidance. Luckily, semantic segmentation maps for urban scenes are ample (at least for the source domain), and methods to generate them on-the-fly are reliable and accurate \textit{ENOUGH} nowadays. Hence, we introduce specialization within the generator architecture --- we split the original image into multiple images of the same size, each such image holds pixels related to specific entities. In our case, these are vehicles only, roads, and everything else (see Figure~\ref{fig:Framework}, generator inputs). In each image, the pixels that do not depict the relevant parts are blacked out. 

Then, instead of a single "pipeline" that is fed with a single image and outputs its translation, we propose duplicating a few of the first layers of our generator, each path for a specific type of entities (see Figure  \ref{fig:MainMultistream}). We call each such path a Stream, making the proposed generator a \textit{Multi-stream} one. 
In this way, each stream addresses only a specific range of entities, encouraging different, or specialized,  translation behavior to the stream. In other words, we integrate the strong semantic segmentation prior through the architecture itself. Of course, after these several chosen duplicated layers, the Streams are merged back together through concatenation, as depicted in Figure \ref{fig:MainMultistream}. This final step helps in informing each stream about the rest of the image and in generating a seamless output. 

\subsection{Semantic Segmentation Regularization Loss}

In addition, we propose adding content preserving losses. A simple one is an edge loss, comparing the edges found in the image before and after the translation. We found this loss helps, as can be expected, with image sharpness, and preservation of structure. To extract the edges, we found a simple isotropic Canny filter to be sufficient \cite{canny}, however this of course could be replaced with other edge encouraging mechanisms.

The second additional loss term we propose employs the in-domain semantic understanding we assume the source domain carries. We observe that if a translation operation indeed preserves the scene structure as it is expected to, its semantic map does not change. In other words, extracting said map from the translation output should yield the same result as the original segmentation, unless the structure has been lost. Hence, we compare the semantic segmentation map of the original image with one generated from the translation result, offering a regularization effect to the translation process. Note that if a segmentation extraction network exists for the target domain, it can be used, but even if one does not exist, it can be trained along with the generator. 

Even though it might seem counter-intuitive at first, we have found this loss to have significant merit. For example, this loss greatly mitigates the aforementioned additional spurious light sources phenomenon. This is the case since blobs of light have the same appearance whether they were falsely added to an empty patch of road or correctly added to the back of a car, which means wrong light sources induce mistakes in the produced segmentation maps. We note that for this purpose, the jointly trained semantic segmenter does not need to be highly accurate, and hence is not required to be powerful and parameter intensive.

\subsection{Hierarchical Generation}

Lastly, we also observe that for complex scenes, the task of generating high resolution details and that of generating the general structure and mode of translation to the different parts of the image, are difficult to learn at the same time, and are actually almost independent. Hence, following recent approaches that demonstrate the benefit or hierarchical generation \cite{rottshaham2019singan}, we propose duplicating the generator twice, in a two step approach. The Multi-stream generation described in Section~\ref{sec:multistream} is performed on a low resolution version of the input and output domains, and another generation step, possibly using the same backbone architecture, is used to added finer-scale details, converting the low-res output of the first step to a full-res one. The architecture for this step in the framework is depicted in Figure~\ref{fig:upscale}. As can be seen by the generator's architecture (Figure \ref{fig:Upsampling}, we do not recommend the Multi-stream approach for this step, but rather incorporating only two paths - the low-res guidance images path, and the full source image one. This is because this step should add more globally aware details, while the low-res guidance image helps with resolving the multi-modal nature of the translation. 

\subsection{Discriminators and Losses}

For the adversarial loss, we also propose employing discriminators that address the different modalities of the data. We employ three different discriminators: a local, low receptive field one; a global, high receptive field one; and a low capacity discriminator with a medium receptive field.

The first two discriminators are able to attend to features in different scales, the latter discriminator is used to stabilize the training and provide consistent gradients to the generators.
For the adversarial loss we use the LSGAN \cite{arxiv1611.04076} term:


\begin{equation}
\label{lsgan}
\begin{split}
 {L}_{LSGAN}(G,D) = \\
 &\frac{1}{2}\mathbb{E}_{\mathbf{y}\sim {p}_{data}(\mathbf{y})}[(D(\mathbf{y}) - {1})^{2}] + \\
 &\frac{1}{2}\mathbb{E}_{\mathbf{x}\sim {p}_{data}(\mathbf{x})}[(D(G(\mathbf{x})))^{2}] + \\
 &\frac{1}{2}\mathbb{E}_{\mathbf{x}\sim {p}_{\mathbf{x}}(\mathbf{x})}[(D(G(\mathbf{x})) - {1})^{2}] \\ \\ 
\end{split}
\end{equation}

And since we employ three different discriminators our adversarial loss is:
\begin{equation}
\label{adv loss}
\begin{split}
L_{Adv} = \\
&\lambda_h L_{LSGAN}\left(G_{X,Y},D_{Y_{high}}\right) + \\
&\lambda_m L_{LSGAN}\left(G_{X,Y},D_{Y_{med}}\right) + \\
&\lambda_l L_{LSGAN}\left(G_{X,Y},D_{Y_{low}}\right)
\end{split}
\end{equation}

In addition, we employ a cycle consistency loss and identity losses, both borrowed from the original employed cycleGAN backbone \citep{CycleGAN2017}:
\\
\begin{equation}
\label{cyc & ind losses}
\begin{split}
&L_{Cyc} = \mathbb{E}_{\mathbf{x} \sim p_{\mathbf{x}}\left(\mathbf{x}\right)}\left\|G_{YX}\left(G_{XY}\left(\mathbf{x}\right)\right) - x\right\|_{1},
\\
&L_{Idn} = \mathbb{E}_{\mathbf{y} \sim p_{\mathbf{y}}\left(\mathbf{y}\right)}\left\|(G_{XY}\left(\mathbf{y})\right) - y\right\|_{1} 
\end{split}
\end{equation}

For the semantic segmentation loss, we train a neural network, based on PSPnet \cite{zhao2017pspnet}, jointly with the rest of the networks to perform semantic segmentation in the translated image.
The 'ground truth' is either provided, or is achieved by the semantic segmentation map (calculated offline) for the original image.
Both the Canny edge loss and the semantic segmentation loss encourage the generator the generate sharper and clearer images, as any random artifacts would create new edges and would limit the semantic segmentation network's ability to accurately predict the correct entity label.
We have observed that the semantic segmentation loss helps to regularize the generator during training.


\begin{equation}
\label{canny & semseg losses}
\begin{split}
&{L}_{Canny} = \mathbb{E}_{\mathbf{x} \sim {p}_{\mathbf{x}}(\mathbf{x})}\|C({G}_{XY}(\mathbf{x})) - C(x)\|_{1},
\\
\\
&{L}_{SemSeg} = -\mathbb{E}_{\mathbf{x} \sim {p}_{\mathbf{x}}(\mathbf{x})}[S(C(x))\cdot\log(S({G}_{XY}(x))) + 
\\ &(1 - S(C(x)))\cdot\log(1 -S({G}_{XY}(x)))] \\ 
\end{split}
\end{equation}

And the total target is:
\\
\begin{equation}
\label{total target}
\begin{split}
L_{Total} = L_{Adv} +
\lambda_{Cyc}L_{Cyc} + \lambda_{Idn}L_{Idn} +\\
\lambda_{Canny}L_{Canny} + \lambda_{SemSeg}L_{SemSeg}
\end{split}
\end{equation}

\begin{figure*}[h!]
    \centering
    \includegraphics[width=16cm]{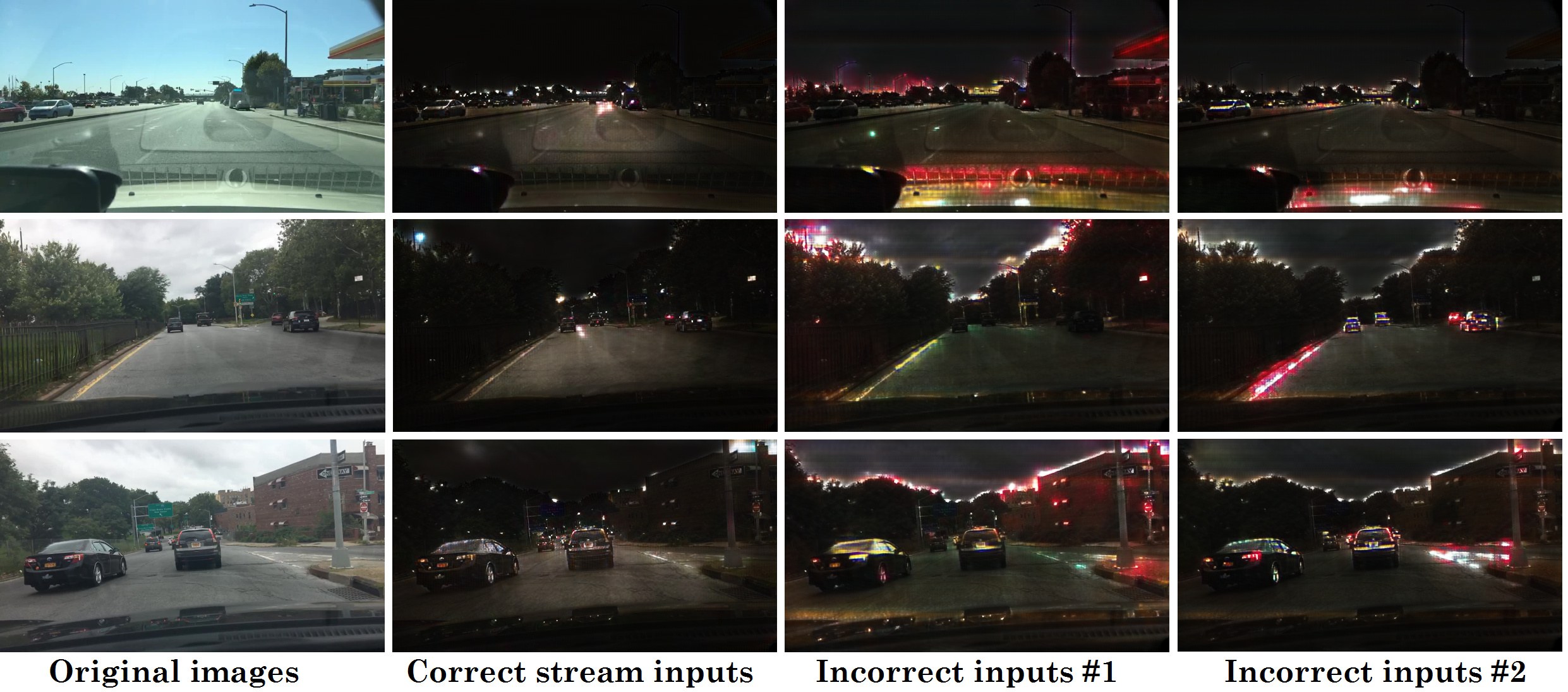}
    \caption{Stream input change. to demonstrate how each stream specialized in translating specific entities, we fed incorrect inputs, i.e., car pixels into a stream that specializes in translating building pixels, and show a degradation in translation quality. Zoomed-in viewing recommended.}
    \label{fig:stream_change}
\end{figure*}


\section{Results and discussion}
\label{sec:results}

\begin{figure*}[h!]
    \centering
    \includegraphics[width=16cm]{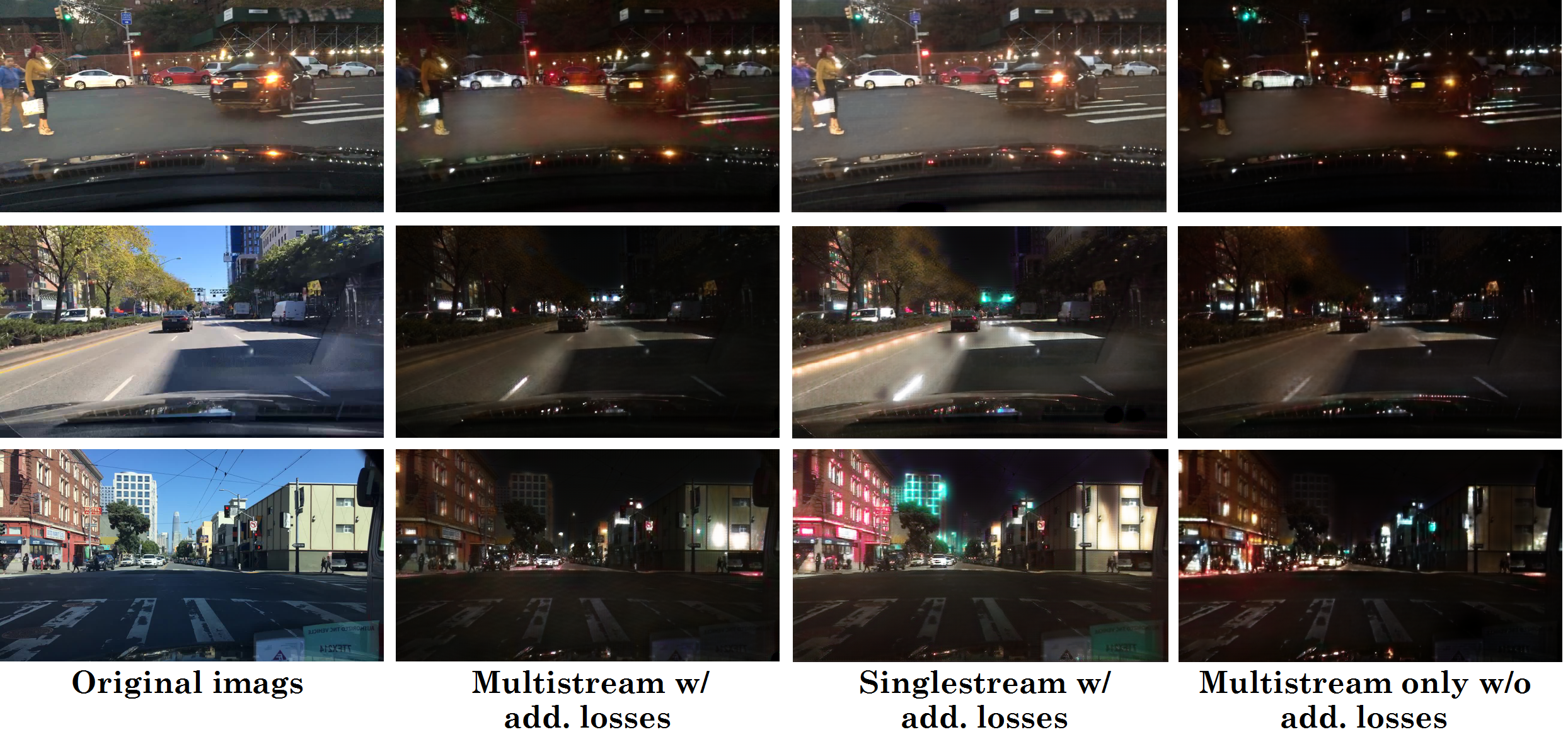}
    \caption{Ablation study. From left to right: input images, the overall proposed framework results Vs. a single-stream generator with all additional losses Vs. a Multi-stream generator without the Canny edge detection, semantic segmentation, global discriminator and low-capacity discriminator losses. As can be seen, the single-stream variant produces results that are more similar to the source domain, and the variant without the rest of the losses produces artifacts such as additional light sources. Zoomed-in viewing recommended. }
    \label{fig:ablation}
\end{figure*}

To demonstrate the power of the Multi-stream generator, we experiment with a challenging task: day-to-night translation using the BDD100K dataset \cite{yu2018bdd100k}.
We chose this dataset as it presents multiple textures and entities, complex geometry and multiple light sources.
Furthermore, we examine an application of our translation for downstream tasks, by  enhancing nighttime detection results achieved using augmented data generated by our Multi-stream generator.

\begin{figure*}[h!]
    \centering
    \includegraphics[width=16cm]{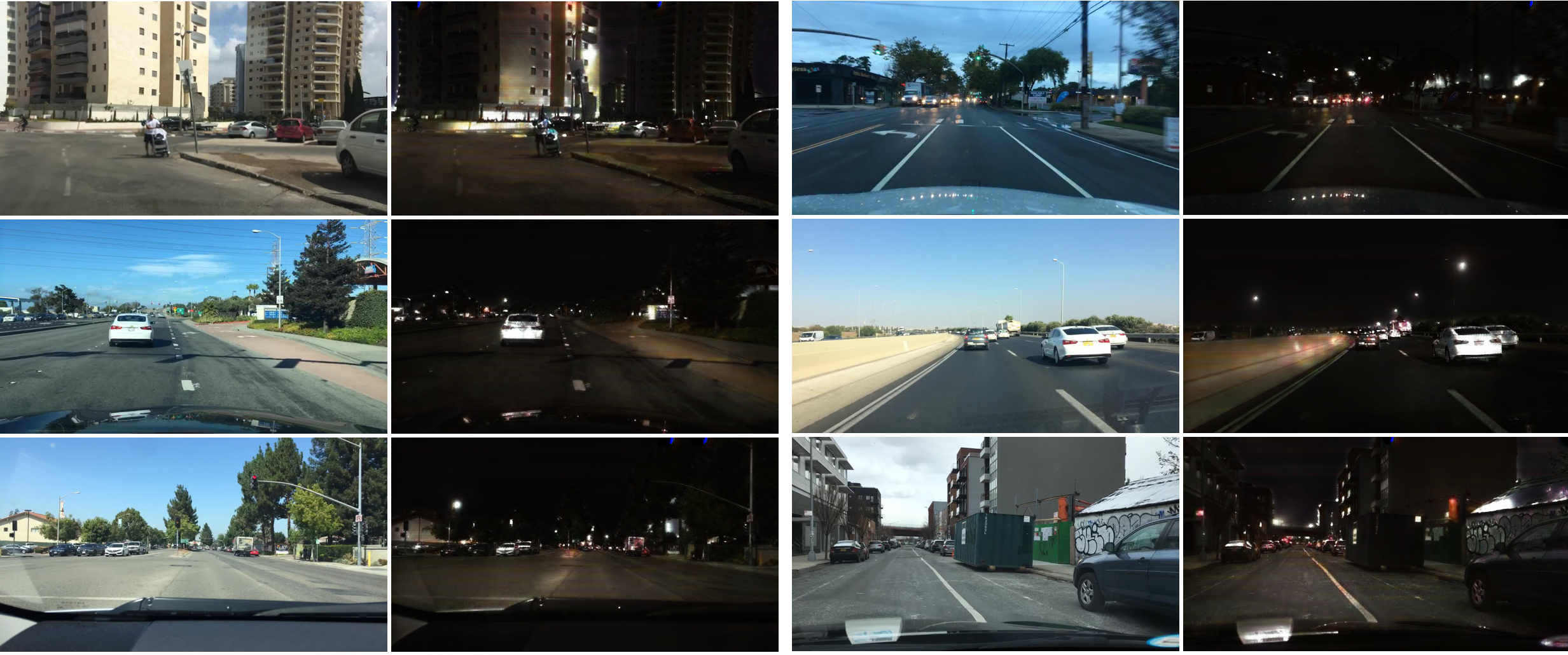}
    \caption{Uncurated day-to-night translation results. Zoomed-in viewing recommended.}
    \label{fig:uncurated}
\end{figure*}

\textbf{For comparison with our model}, we show baseline results using the CycleGAN~\cite{CycleGAN2017}, MUNIT~\cite{huang2018munit}, and FUNIT~\cite{liu2019few} models, which are the pillars of UI2I state-of-the-art.

As can be seen in Figure~\ref{fig:comparisons}, Single-stream models struggle with the translation of complex scenes, due to the different translation distributions the different semantic regions of the images present. This manifests in two typical modes of failure. The first is partial translation, yielding images that still keep some of the daytime features. This is especially noticeable in the CycleGAN results, and in the sky regions of the MUNIT ones.

More over, single-stream models tend to produce noticeably more artifacts, depicting translation modes typical to one region in another. The most prominent example is the addition of spurious light sources, scattered almost randomly through the resulting images.  

\textbf{Stream specialization}. We provide evidence that enforcing the semantic segmentation prior within the generator's architecture itself indeed produce streams that are \textit{specialized}, i.e., that learn a translation distribution that is typical to a specific entity type. To do this, we have tested our trained generator using incorrect semantic segmentation maps.
For each generator Stream we switch input maps, containing entities it was not trained for.
If translating each entity does not require different processing, and if each Stream did not have to specialize in translating specific entities, providing "incorrect" inputs (meaning images containing entities it was not trained upon) would not result in any meaningful or noticeable degradation in translation quality. As can be seen in Figure~\ref{fig:stream_change}, this mixup heavily affects results quality, as for example, feeding roads to the stream specializing in cars yields headlights and backlights on road markings.


\textbf{Downstream Application}
We perform a detection task on the BDD100K dataset using the YOLOv5 model \cite{glenn_jocher_2021_4418161}, using our translation model as an augmentation method. For this, we train two YOLOv5 models, one trained on the original dataset (real night + real day images), while the other is trained on the original night images, and translated day-to-night ones (real night + fake night) images.  
We test both models on real night images that were not used during training.
The class labels we have used were Person, Car, Bus and Truck, we chose these classes as they represent non-background objects, as opposed to the sky, buildings or vegetation.
As can be seen in the following table, the YOLOv5 detection model achieved better results using our translated fake night images, thus providing evidence for the quality of our models outputs. 

\begin{table}[t]
\caption{Classification accuracies of YOLOv5 using real night + real day data (left), and real night + fake night data translated by our method (right).}
\label{sample-table}
\vskip 0.15in
\begin{center}
\begin{small}
\begin{sc}
\begin{tabular}{lcccr}
\toprule
Class & mAP@.5 &  mAP@.5:.95 & Recall \\
\midrule
Person    & \textbf{0.405} / 0.402 & 0.16 / \textbf{0.161} & 0.507 / \textbf{0.520} \\

Car   & 0.677 / \textbf{0.681} & 0.361 / \textbf{0.362} & 0.754 / \textbf{0.758} \\

Bus   & 0.477 / \textbf{0.496} & 0.337 / \textbf{0.360} & 0.563 / \textbf{0.593} \\

Truck   & 0.508 / \textbf{0.519} & 0.332 / \textbf{0.338} & 0.556 / \textbf{0.593} \\

\bottomrule
\end{tabular}
\end{sc}
\end{small}
\end{center}
\vskip -0.1in
\end{table}

\subsection{Ablation study}

To justify our design choices, we compare the results of our framework against several other alternatives. two experiments can be found in Figure~\ref{fig:ablation}, while other experiments, investigating other alternative architectures and losses, can be found in the Supplementary material. 
In the first experiment, we a single-Stream architecture, keeping all the rest of the proposed framework intact.
In the second, we train a Multi-stream generator, but we omit the rest of the proposed framework, i.e. the semantic segmentation loss, canny edge loss and the global high receptive field and low-capacity medium receptive field discriminators.

As can be observed in the ablation results in \ref{fig:ablation}, the Single-stream generator struggles to perform the full nighttime conversion, as its ability to process correct translation for each entity is more limited.
Also, the Multi-stream generator, without the additional loss terms, does produce convincing darkening effects, but the generated images are blurrier and contain less detail.
As for computation resources, all of our experiments were conducted on twin GeForce RTX 2080 Ti 11GB, training the Multi-stream network required 5 days of computation time while training the Upscaling network required 2 days of computation time.


\section{Conclusions}
\label{sec:discussion}
We have shown that in-domain knowledge, such as semantic segmentation, even only in the source domain, can be leveraged to improve UI2I tasks, especially for complex scenes. In such scenes different types of entities induce completely different distributions. We have demonstrated that learning the modes of the translation is already a difficult task, and hence hinting the network which region in the image corresponds to which type of translation reduces the load off the optimization, and provides powerful guidance. 

In this paper, we have demonstrated the applicability of our method on urban scenes, but our method is not restricted to these regime alone. It could of course be also applicable to many other multi-modal scenarios, such as landscapes, room scenes, etc. Using these scenes, the method could be directly applied to translation tasks between domains of different weather conditions (such as winter, summer, haze, etc.) or appearances (such as paintings, IR, etc.).

Moreover, we have demonstrated the method by applying our introduced adaptation to the CycleGAN architecture. However, our Multi-stream framework is agnostic to the backbone generator's architecture. Hence, applying it to other architectures, could yield even more merit, and is an interesting direction to investigate further. 

Looking into the future, we see several paths of interesting research. Firstly, we have seen that the day-to-night translation results of our method are convincing, but we can also see that our method is not able to cope with shadows well, as shadows in the input images still appear in the output ones. Investigating and integrating a shadow removal technique into the Multi-stream framework could prove very beneficial, perhaps integrating some long-distance relationship architectures, such as transformers.

Taking a different approach, the proposed framework, with some investigation, could also be leveraged without explicit in-domain supervision. The sole assumption that the translation task should be decomposed into different regions that bear semantic and translation behavior similarities could already be enough to empower a self-supervised method that searches for the said decomposition along with the translation itself. Such an approach could, for example, employ clustering or Reinforcement Learning techniques. 

Finally, we hope to see in the future more works that leverage the understanding of a some data in order to help a task over a larger set. More specifically, we believe that in-domain supervision could be manifested in more ways than semantic segmentation, and expect to witness further investigation in this direction, incorporating more insights of one domain into the other, in the near future.  


\section{Appendix}\label{sec:Appendix}
\begin{figure*}[t]
    \centering
    \includegraphics[width=16cm]{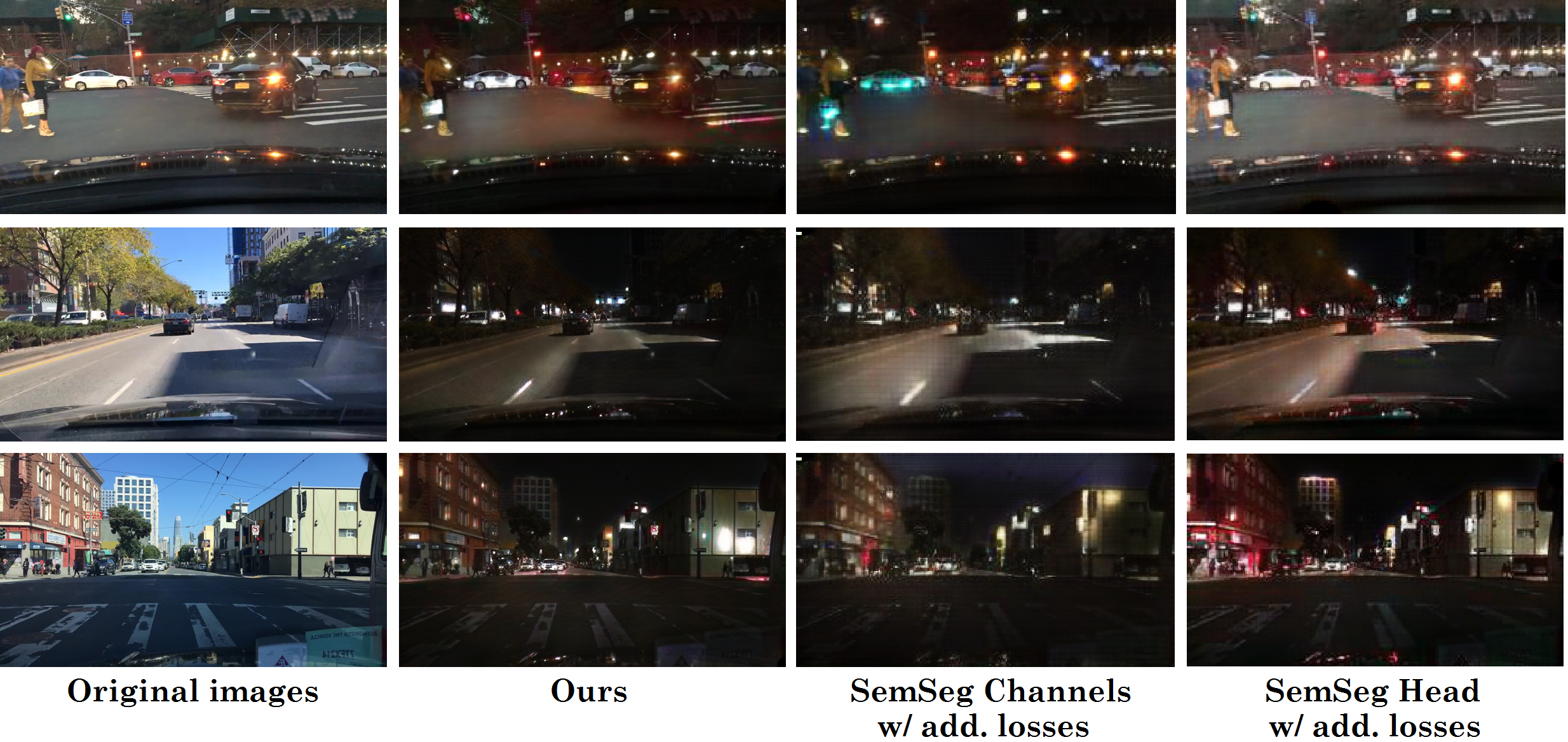}
    \caption{Different methods of using the semantic segmentation prior: Multi-stream w/ add. losses (ours), concatenating the semantic segmentation map as an extra channels in a single-stream generator w/ add. losses and a single-stream generator which outputs a semantic segmentation prediction of the input w/ add. losses}
    \label{fig:semsegchannelsandhead}
\end{figure*}

We wish to demonstrate the effectiveness of our method of leveraging the semantic segmentation prior, in our method, the semantic segmentation prior is being utilized fundamentally in the architecture via Multi-streaming, but other methods of utilizing it exist.
One method is providing extra channels as input to the generator (on top of the RGB), where the extra channels contain the semantic segmentation map (dubbed as SemSeg Channels), the other method is adding a semantic segmentation head to the generator, where it is tasked the generate the semantic segmentation map of the input image.
We have embedded these generators in our framework and pitted them against our method.
As shown in Figure~\ref{fig:semsegchannelsandhead}, these method do benefit from the additional prior knowledge, but fail to provide the translation quality of the Multi-stream generator.
For both of the compared generators, we have  \textit{significantly} increased the number of the parameters, surpassing that of the Multi-stream generator by over 40\%.

\subsection{Implementation Details}
\label{sec:impdet}
Our Up-scaling network receives the original image as well as a low-res guidance image to aid in the translation process, this low-res guidance image is a translated low-res image generated by the Multi-stream generator, the task of the Up-scaling network is to add fine details to provide a high-res translation.
During training, in order to encourage the Up-scaling network to make use of the guidance image, when reconstructing the generated fake image from the inverse generator (original domain B image-> fake domain A image -> reconstructed original) , we input the blurred original image as the guidance image.
This encourages the generator rely on the guidance image for translation, essentially removing this heavy burden from its shoulders, and encourages it to focus on adding fine details, resulting in a high-res translation. 

\bibliographystyle{unsrtnat}
\bibliography{references} 

\end{document}